\documentclass{article}

\usepackage{graphicx}
\usepackage{wrapfig}
\usepackage{enumitem}
\usepackage{amsmath}

\usepackage[nonatbib, final]{neurips_bdl2019}

\usepackage[utf8]{inputenc} 
\usepackage[T1]{fontenc}    
\usepackage{hyperref}       
\usepackage{url}            
\usepackage{booktabs}       
\usepackage{amsfonts}       
\usepackage{nicefrac}       
\usepackage{microtype}      
\usepackage{xcolor}
\usepackage{cite}

\title{The Spectral Bias of the Deep Image Prior}

\author{
  Prithvijit Chakrabarty 
  \thanks{Currently at Amazon. This is work done as a graduate student at the University of Massachusetts}
  \quad 
  Subhransu Maji \\
  College of Information and Computer Sciences \\
  University of Massachusetts, Amherst\\
  \{\tt\small pchakrabarty,smaji\}@cs.umass.edu
}

\begin{document}

\maketitle



\begin{abstract}
   The ``deep image prior'' proposed by Ulyanov et
   al. \cite{ulyanov2018deep} is an intriguing property of neural
   nets: a convolutional encoder-decoder network can be used as a prior for
   natural images. 
   The network architecture implicitly introduces a bias; If we train
   the model to map white noise to a corrupted image, this
   bias guides the model to fit the true image before fitting the
   corrupted regions.
   This paper explores \emph{why} the deep image prior helps in denoising natural
   images. 
   We present a novel method to analyze trajectories generated by the deep image prior optimization and demonstrate:
   (i) convolution layers of the an encoder-decoder decouple
   the frequency components of the image, learning each at different
   rates (ii) the model fits lower frequencies first, making early stopping behave as a low pass filter.
   The experiments study an extension of Cheng
   et al.~\cite{cheng2019bayesian}, which showed that at
   initialization the deep image prior is equivalent to a stationary
   Gaussian process.
\end{abstract}

\section{Introduction}
It is well known that large neural nets have the capacity to overfit
training data, even fitting random
labels perfectly~\cite{zhang2016understanding}.
Arpit et al.~\cite{arpit2017closer} confirmed this, but showed
that networks learn "simple patterns" first. 
This may explain why these models, despite their capacity, generalize
well with early stopping.
\cite{neyshabur2017exploring,gunasekar2018implicit,soudry2018implicit,neyshabur2014search}.
One way to formalize
the notion of a simple pattern is to observe the frequency
components of the function learned by the model. 
Simple patterns are smooth, composed of low frequencies. 
A number of works use this approach to describe the bias induced by deep networks. 
For example,
\cite{xu2018training,xu2019frequency} study the
\textit{F-principle} 
which states that the
models first learn the low-frequency components when fitting a signal.
Rahaman et al.~\cite{rahaman2018spectral} demonstrated a similar
\textit{spectral bias} of a deep fully-connected network with ReLU
activation. 

The Deep Image Prior (DIP) considers the following
setup. 
Let $f_\theta:\mathcal{X} \to \mathcal{X'} $ be a convolutional
encoder-decoder parameterized by $\theta \in \Theta$. $\mathcal{X}$, $\mathcal{X'}$
are spaces of $D$ dimensional signals. 
DIP studies the following optimization:
$
    \min\limits_{\theta} ||f_\theta(z)-x_0||^2
$
where, $z \in \mathcal{X}$ is a fixed $D$ dimensional white noise
vector. 
$N$ steps of gradient descent for this optimization traces out a trajectory
in the parameter space: 
$
\theta^{(1)},
... \theta^{(N)}
$. This has a corresponding trajectory $T$ in the
output space $\mathcal{X'}$:
$
f_{\theta^{(1)}}(z), ...,
f_{\theta^{N}}(z)
$. 
Given enough capacity and suitable learning rate scheme, for a large $N$, the model will perfectly fit the
signal, i.e., $f_{\theta^{N}}(z) = x_0$. 
For image denoising $\mathcal{X}$, $\mathcal{X'}$ are
spaces of all images ($D=2$) and $x_0 = x^* + \eta$ is a noisy image: a
clean image $x^*$ with added Gaussian noise $\eta$. Experiments in
\cite{ulyanov2018deep} show that early stopping with gradient descent will lead to denoised
the image. 
In other words, the trajectory $T$ will contain a point that is close to the clean image $x^*$.

We use the spectral bias of the network to explain
this denoising behavior.
It is known that at initialization the generated output of the DIP is drawn from distribution that is approximately a stationary Gaussian process with smooth covariance function~\cite{cheng2019bayesian}. 
The experiments presented here suggest that
this trend continues throughout the optimization, i.e., the model
learns to construct the image from low to high frequencies. 
Thus, early stopping prevents fitting the high frequency components introduced
by the additive Gaussian noise. The source code to reproduce results is available online\footnote{\url{https://github.com/PCJohn/dip-spectral}}.

\begin{figure*}
\newcommand{\hh}{45pt}
\newcommand{\ww}{45pt}
\centering
\setlength{\tabcolsep}{1pt}
\begin{tabular}{ c c c c c c c c}
\includegraphics[width=\ww,height=\hh]{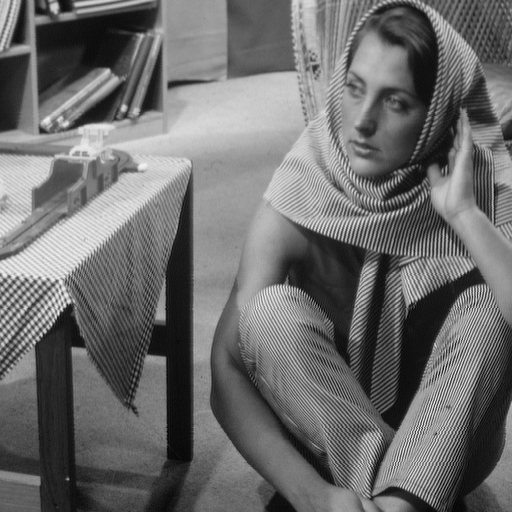} &
\includegraphics[width=\ww,height=\hh]{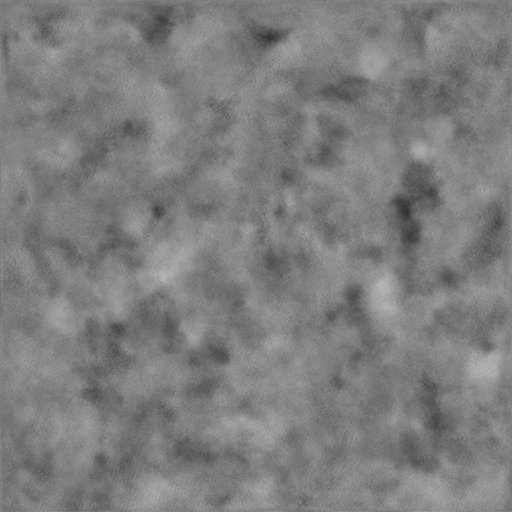} &
\includegraphics[width=\ww,height=\hh]{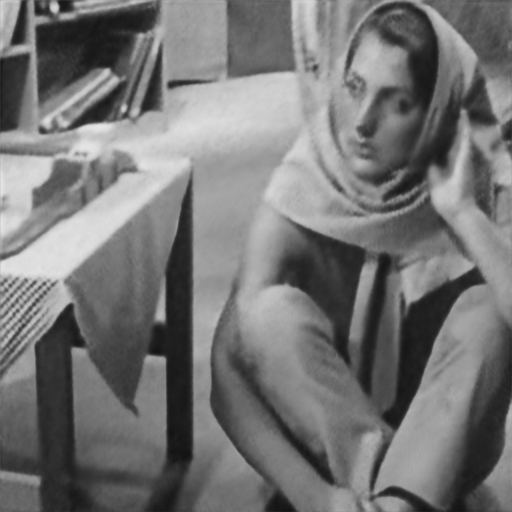} &
\includegraphics[width=\ww,height=\hh]{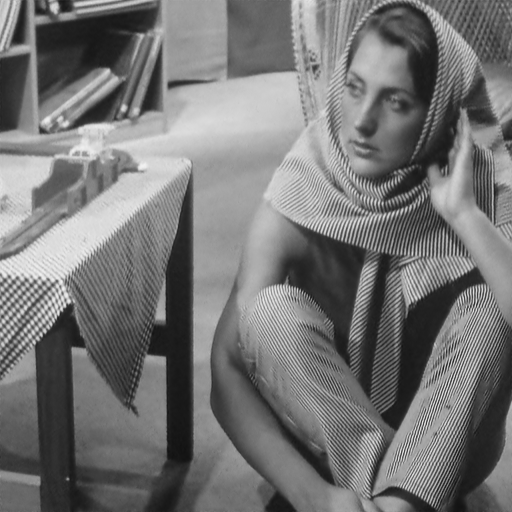} &
\includegraphics[width=55pt,height=\hh]{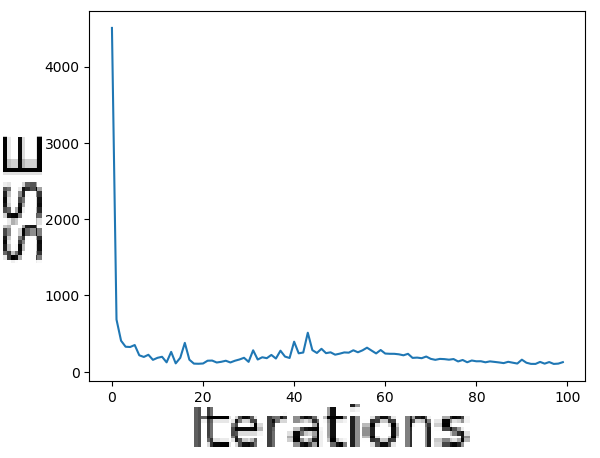}& 
\includegraphics[width=\ww,height=\hh]{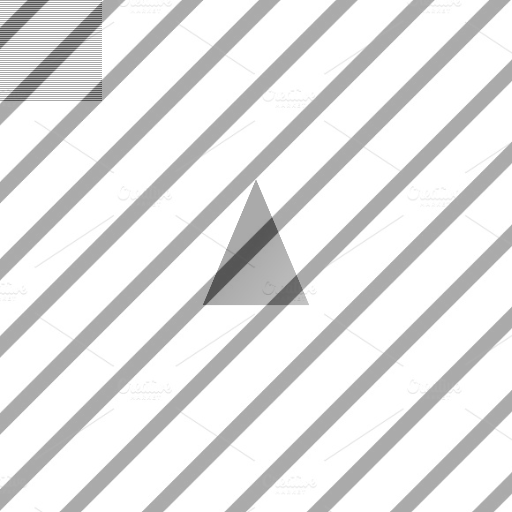} &
\includegraphics[width=\ww,height=\hh]{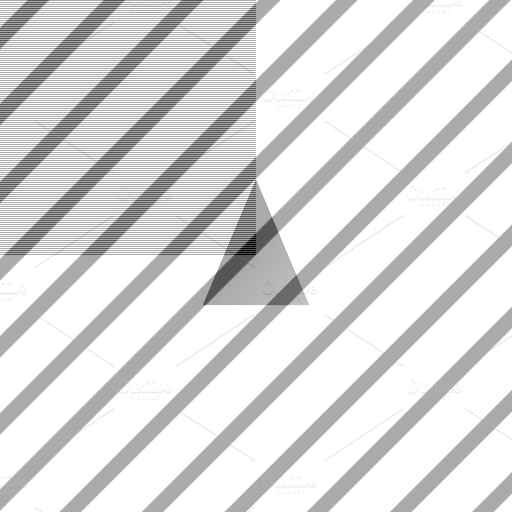} &
\includegraphics[width=\ww,height=\hh]{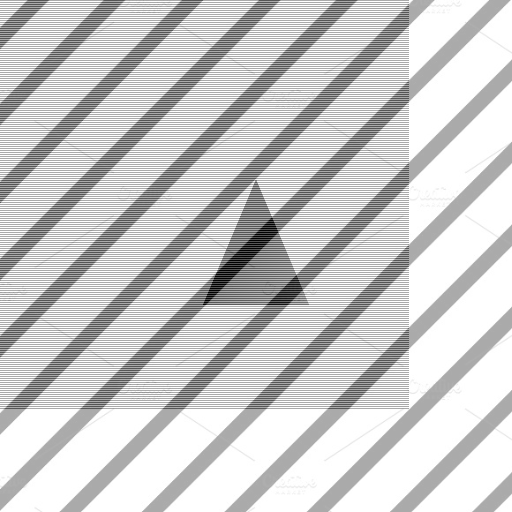} \\

\includegraphics[width=\ww,height=\hh]{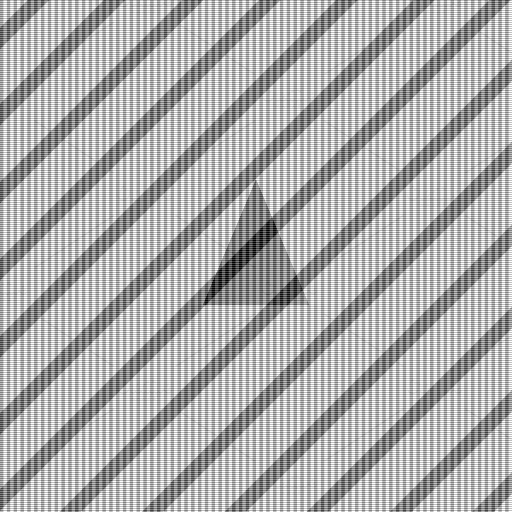} &
\includegraphics[width=\ww,height=\hh]{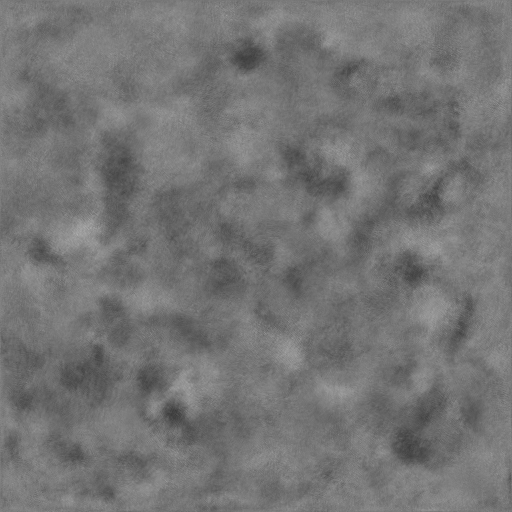} &
\includegraphics[width=\ww,height=\hh]{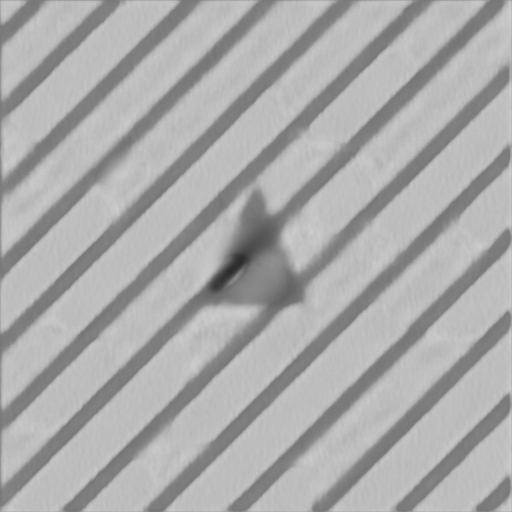} &
\includegraphics[width=\ww,height=\hh]{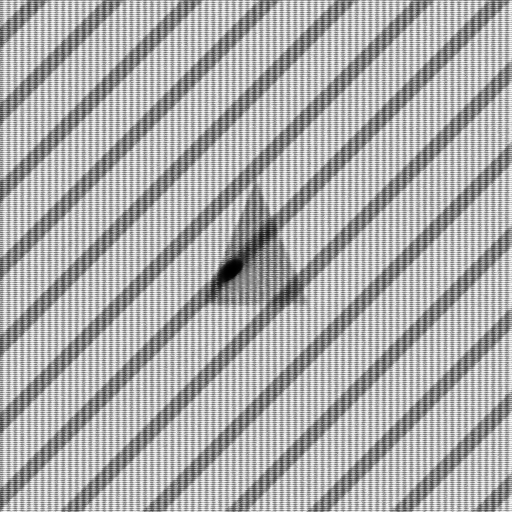} &
\includegraphics[width=55pt,height=\hh]{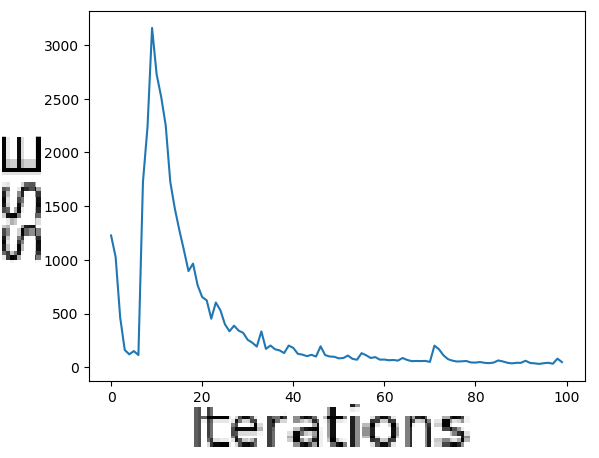}& 
\includegraphics[width=\ww,height=\hh]{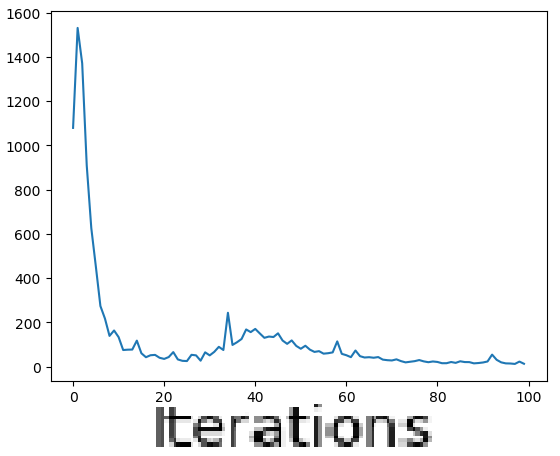} & 
\includegraphics[width=\ww,height=\hh]{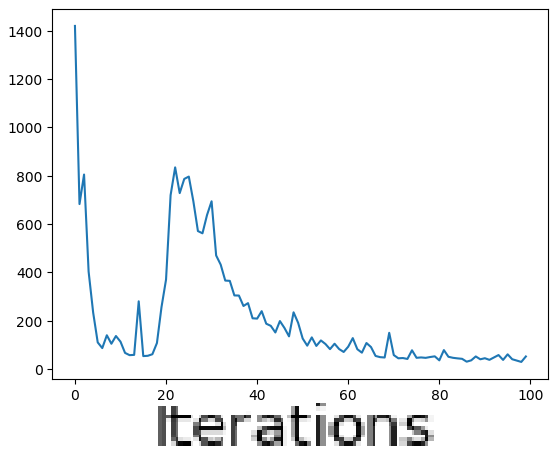} &
\includegraphics[width=\ww,height=\hh]{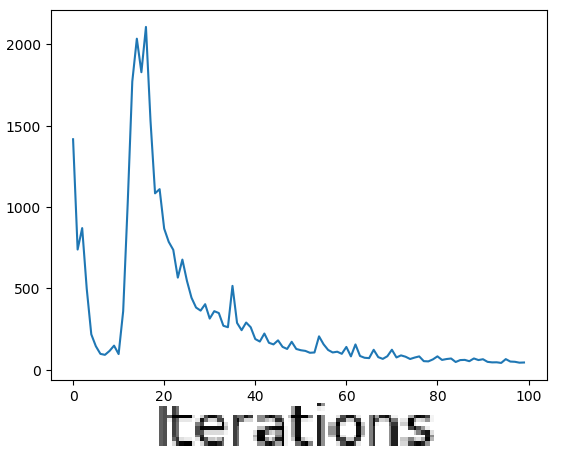} \\
(a) & (b) & (c) & (d) & (e) & (f) & (g) & (h)\\
\end{tabular}
\caption{\textbf{(a-e) DIP applied to two \textit{clean} images.} (a) Input images. (b)
  Initial output. (c) Output at the first convergence of trajectories.
  (d) Output at the second convergence of trajectories. (e) Variation
  of $\varepsilon$ across iterations. 
  \textbf{(f-g)} Controlled amounts of high frequency patterns on
  triangle.png. Trajectories diverge more when the pattern covers a
  larger area. \emph{(Please zoom in for details.)}}
\label{dip_demo}
\end{figure*}

\section{The Spectral Bias of the Deep Image Prior}
We first demonstrate the spectral bias of the DIP in the exact setting as
Ulyanov et al.~\cite{ulyanov2018deep}. 
Using the same model, i.e., a 5-layered
convolutional autoencoder with $128$ channels each, the DIP
optimization was run on
\textit{clean} images which have a range of frequency components, as
shown in Figure~\ref{dip_demo}(a). 
For each image, the optimization was run
twice to generate trajectories $T_1$, $T_2$ in the output space.
Let $\mathcal{\varepsilon}$ be the variation of the sum of squared error
(SSE) between trajectories: $\mathcal{\varepsilon}^{(i)} = \sum
\big(T_1^{(i)}-T_2^{(i)}\big)^{2}$.

\begin{wrapfigure}{r}{0.5\textwidth}
\centering
\vspace{-10pt}
\newcommand{\hh}{65pt}
\newcommand{\ww}{65pt}
\setlength{\tabcolsep}{1pt}
\begin{tabular}{ c c c }
\includegraphics[width=\ww,height=\hh]{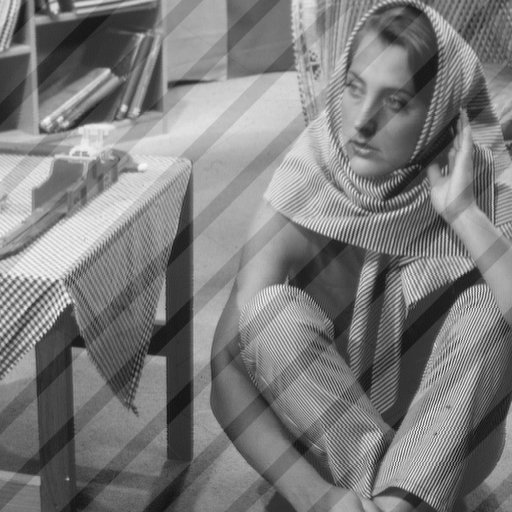} &
\includegraphics[width=\ww,height=\hh]{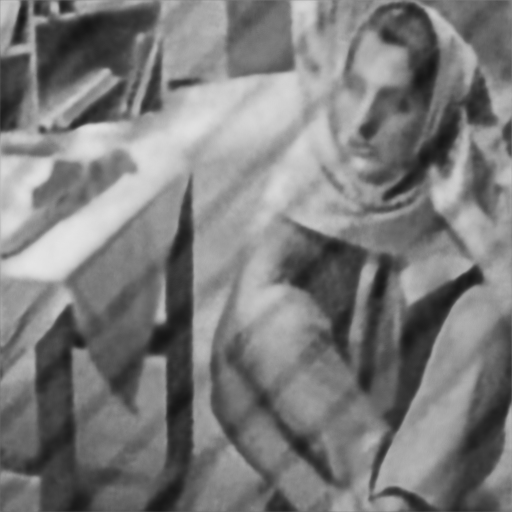} &
\includegraphics[width=\ww,height=\hh]{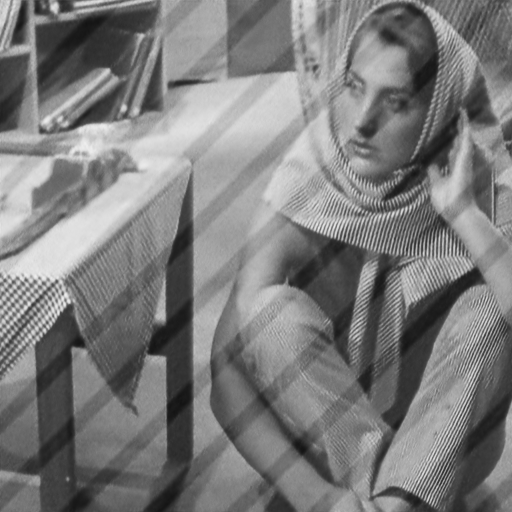} \\
\includegraphics[width=\ww,height=\hh]{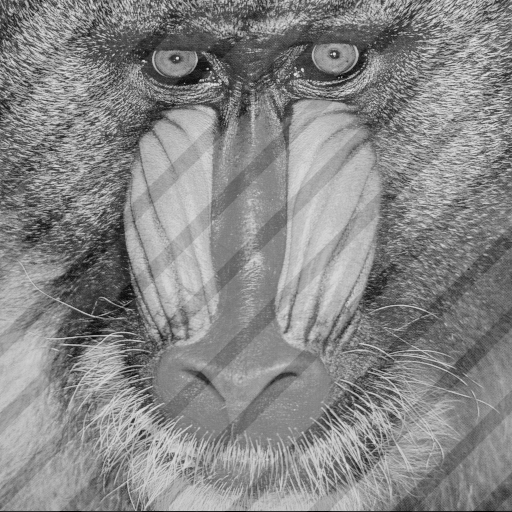} & 
\includegraphics[width=\ww,height=\hh]{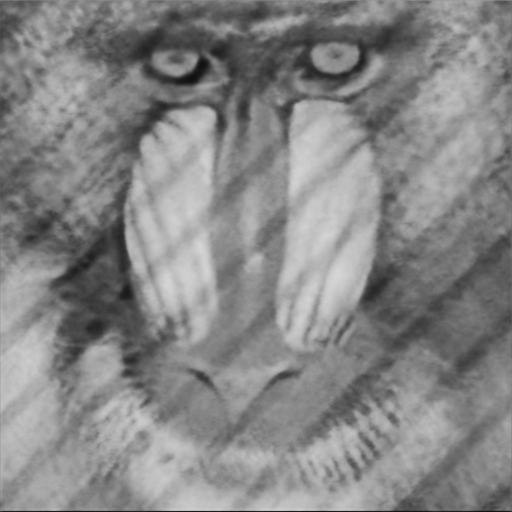} &
\includegraphics[width=\ww,height=\hh]{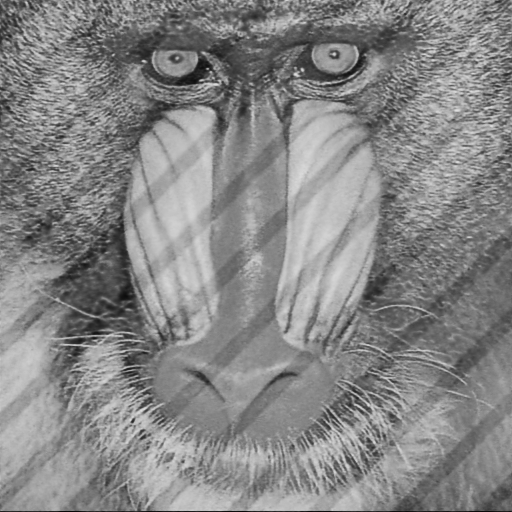} \\
(a) & (b) & (c)
\end{tabular}
\caption{\textbf{Some failure cases for the DIP.} (a) Images with low frequency noise. (b) $20$ iterations: the model fits the noise, not the true high frequency components. (c) $500$ iterations: the model fits the signal perfectly.}
\label{dip_fail}
\end{wrapfigure}

Fig.~\ref{dip_demo} tracks the $\mathcal{\varepsilon}$ on two images: barbara.png
naturally has high frequency components while triangle.png is a smooth
image with a high frequency pattern superimposed on it.
We observe the following:
\begin{itemize}[label=\textbullet,nosep,leftmargin=*]
\item The initial output of the model is smooth
  (Fig.~\ref{dip_demo}b). This is consistent with the analysis in
  \cite{cheng2019bayesian}, which showed that at initialization the
  output can be approximated as a draw from a stationary Gaussian
  process with a smooth covariance function.
\item The trajectories converge twice (Fig.~\ref{dip_demo}e). At the
  first point of convergence, the model predicts a low frequency
  reconstruction of the image (Fig. \ref{dip_demo}c). After the
  second converge, the model fits the high frequency components
  (Fig.~\ref{dip_demo}d).
\item Controlled amounts of the high frequency pattern were added to
  triangle.png. As it covers a larger spatial extent, the trajectories
  diverge more (Fig. \ref{dip_demo}(f-h)). This suggests that the
  reason for divergence corresponds to the model learning the higher frequencies.
\end{itemize}

These observations show that the model learns the frequency components
of the image at different rates, fitting the low frequencies first
(see Appendix \ref{app_power_spec}). Thus, early stopping is similar
to low-pass filtering in the frequency domain.

The denoising experiments in \cite{ulyanov2018deep} used additive Gaussian noise. Early stopping with DIP prevents fitting these, predicting the
clean image before fitting the noise. We can now construct samples where deep image prior is guaranteed to
fail. Low frequency noise is added to barbara.png and baboon.png:
images which naturally have high frequencies, as shown in
Fig.~\ref{dip_fail}. 
After $20$ iterations, the model fits the noise, but \textit{not} the
high frequency components and goes on the fit the
input perfectly after $500$ iterations. 
Thus, there is no point at which the model predicts the clean image. 
This strongly suggests that the ability of DIP to denoise images is brought about
due to the frequency bias.


\section{What causes the frequency bias?}

Here we investigate what elements of the DIP lead to the
aforementioned frequency bias in learning. In particular we show that
both convolutions and upsampling introduce a bias.

\newcommand{\hh}{70pt}
\newcommand{\ww}{80pt}
\begin{figure}[t]
\centering
\setlength{\tabcolsep}{1pt}
\begin{tabular}{cccccccc}
\includegraphics[width=90pt,height=\hh]{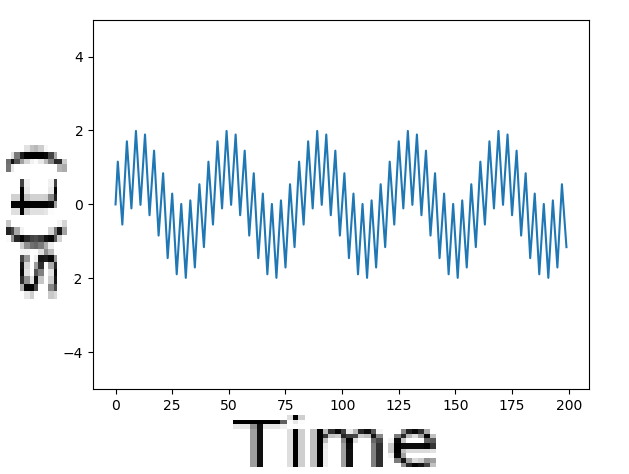} &
\includegraphics[width=\ww,height=\hh]{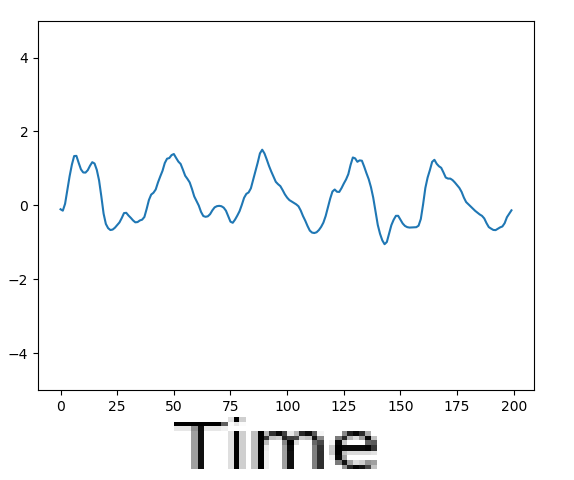} &
\includegraphics[width=\ww,height=\hh]{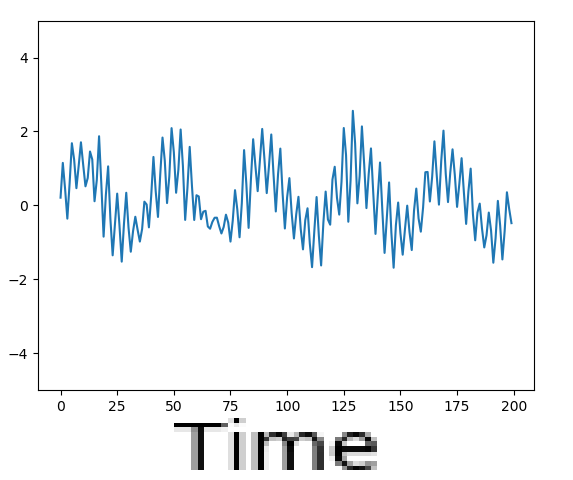} &
\includegraphics[width=\ww,height=\hh]{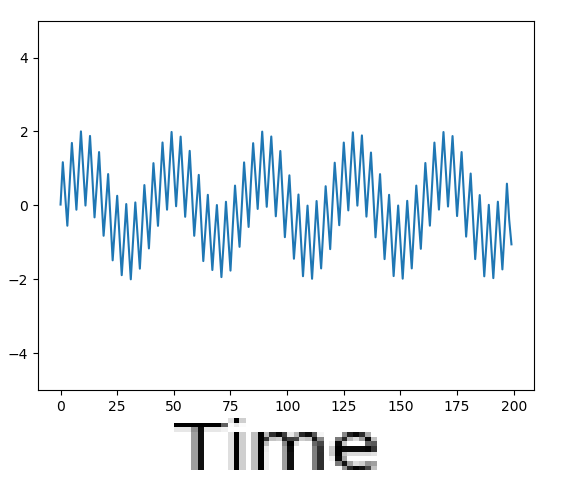} \\
(a) & (b) & (c) & (d) & \\
\includegraphics[width=90pt,height=\hh]{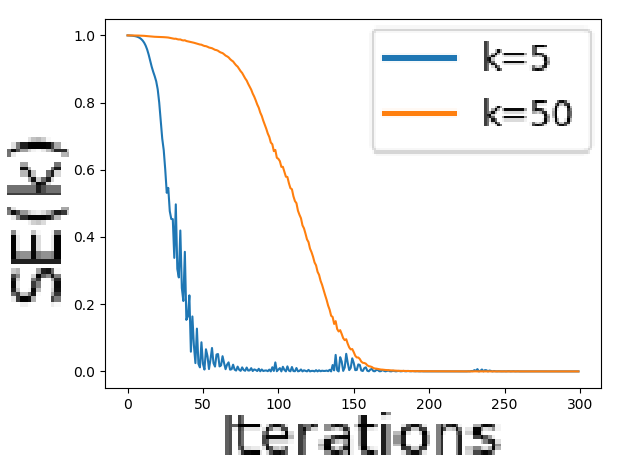} &
\includegraphics[width=\ww,height=\hh]{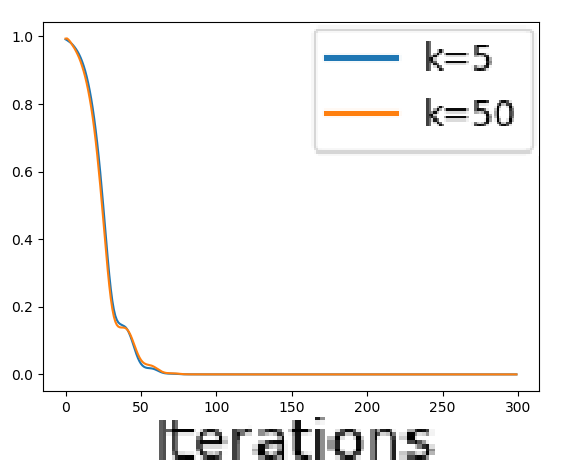} &
\includegraphics[width=\ww,height=\hh]{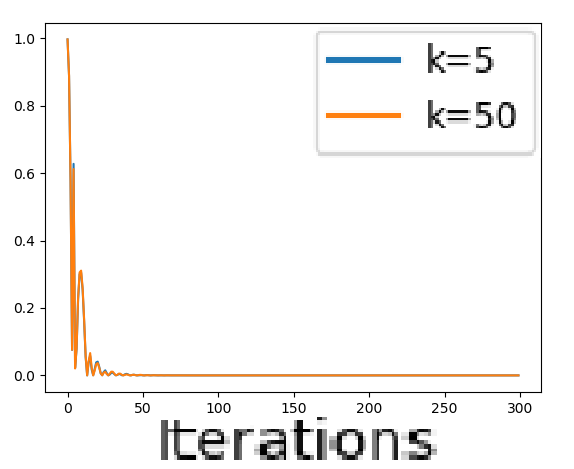} & 
\includegraphics[width=\ww,height=\hh]{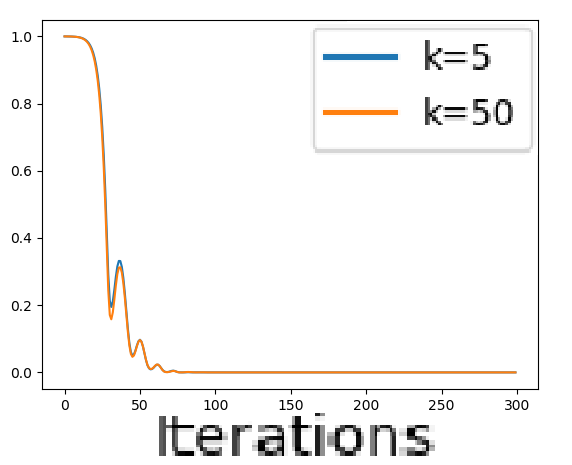} \\
(e) & (f) & (g) & (h)
\end{tabular}
\caption{\textbf{DIP on 1D signals.} \textbf{(a-e)} Outputs of a model
  with conv layers. (a) Input signal (b) Predicted signal when
  $k_1$ converges, (c) when $k_2$
  converges and (d) at the end of the optimization.
  (e) Variation of squared errors
  $SE(k_1)$ and $SE(k_2)$ across iterations. 
\textbf{(f-h)} Outputs of a model with linear layers. Variation of
$SE(k_1)$ and $SE(k_2)$ with (f) 256 nodes per layers (g) 4096 nodes
per layers (h) depth 24. This model fails to
decouple frequencies.}
\label{oned_res}
\end{figure}

\subsection{Convolution Layers} 
\label{conv_layers_bias}
The frequency selectiveness of a convolutional network architecture
was shown in \cite{saxe2011random}.
DIP uses a convolutional encoder-decoder.
Here, we demonstrate using a simple experiment 
that the convolution layers of this architecture decouple the frequencies of the signal, 
fitting each at different rates.
This does not happen 
if the encoder-decoder has only
linear layers.
Consider DIP on 1D signals, where $x_0 = \sin(2\pi k_1 t) + \sin(2\pi
k_2 t)$. 
The signal with $k_1=5$ and $k_2=50$ is shown in fig. \ref{oned_res}(a).
We run the DIP optimization on this signal and track the squared error $SE(k_i) = (\tilde{A_i}-A_i)^2$, where $\tilde{A_i}$ is the predicted amplitude and $A_i$ is the true amplitude for frequency $k_i$.
We say that a frequency $k_i$ has converged at $t_i$ if $SE(k_i) \leq
\delta$ after $t_i$ iterations ($\delta=0.01$ for these
experiments). This is similar to Experiment 1 in
\cite{rahaman2018spectral}. We compare the use of convolutions against linear layers in the model:

\begin{itemize}[label=\textbullet,nosep,leftmargin=*]
\item \noindent\textbf{DIP-Conv.} A 10-layered encoder-decoder with 1D convolution
layers, 256 channels. The variation in $SE(k_1)$ and
$SE(k_2)$ is shown in Fig.~\ref{oned_res}(e). $SE(k_1)$ drops sharply,
leading $k_1$ to converge at 45 iterations. $k_2$ is learned slowly,
converging at 151 iterations. Observing the predictions after each
component converges (Fig.~\ref{oned_res}(b,c)), we see when $k_1$
converges, the model predicts a smooth reconstruction of the
signal. Clearly, the frequency components are learned at different
rates.

\item \noindent\textbf{DIP-Linear.} The 
convolution layers of the DIP-conv were replaced with 256
unit linear layers (fully-connected) and the optimization was run
again. Fig.~\ref{oned_res}(f) shows the results. 
The error for both frequencies drop at the same rate. The model never
predicts a low-frequency reconstruction of the signal. Further, the
lack of decoupling remains if we change the depth or width of the
network (Fig. \ref{oned_res}(g,h)) which suggests that it is intrinsic
to linear layers and not related to the model capacity.
\end{itemize}

\begin{wraptable}{r}{0.5\textwidth}
  \centering
  \resizebox{0.4\textwidth}{!}{%
  \setlength{\tabcolsep}{1pt}
  \renewcommand{\arraystretch}{0.9}
  \begin{tabular}{c  c  c  c}
  
    \hline
    \small{\textbf{DIP}} & 
    \begin{tabular}{cc} 
        \small{\textbf{DIP}} \\ 
        \small{\textbf{Linear-128}}
    \end{tabular} & 
    \begin{tabular}{cc} 
        \small{\textbf{DIP}} \\ 
        \small{\textbf{Linear-2048}}
    \end{tabular} & 
    \small{\textbf{ReLUNet}} \\ 
    \hline
    
    \small{27.47} & \small{20.54} & \small{19.17} & \small{27.58} \\
    
    \hline
  \end{tabular}
  }
  \caption{\textbf{Denoising performance}. Only models with frequency bias are effective at denoising.}
  \label{deniose_res}
\end{wraptable}

\noindent\textbf{Effect on image denoising.}
To confirm that the above results extend to 2D signals, the same architecture variations were applied to images.
\textbf{DIP}: Standard DIP for images using the model from \cite{ulyanov2018deep} (similar to DIP-Conv above);
\textbf{DIP Linear-128}: An encoder-decoder with 5 fully-connected layers, $128$ units each;
\textbf{DIP Linear-2048}: DIP Linear, with $2048$ units each per layer (higher capacity);
\textbf{ReLUNet}: A 10-layered fully-connected network with $256$ nodes per layer and ReLU activations to model images as signals. The model is trained to map pixel coordinates to the corresponding intensities (this will exhibit frequency bias as shown in \cite{rahaman2018spectral}).

The ReLUNet and DIP-Linear architectures do not explicitly have convolution layers. However, their behavior is closely related to special cases of the DIP model. A ReLUNet predicts individual pixels without using neighborhood information, similar to a convolutional encoder-decoder with a kernel size of 1. DIP-Linear is similar to the other extreme, when the kernel size equals the size of the entire signal.


These models were used to fit images used in
\cite{ulyanov2018deep,cheng2019bayesian} from a standard
dataset\footnote{\url{http://www.cs.tut.fi/~foi/GCF-BM3D/index.html}}. 
For efficiency, the images were downsampled by a factor of $4$. 
The entire trajectories in the output space were
saved to see if they contain the denoised image (samples in Appendix
\ref{app_traj_denoise}). To eliminate the effect of a badly chosen
stopping time, we track the PSNR across the entire trajectory and
record the best.

Table \ref{deniose_res} shows the mean PSNR across 9 images (See appendix \ref{per_image_results} for results per image). DIP performs the
best, with ReLUNets achieving comparable performance. Encoder-decoders with only
fully-connected layers perform significantly worse, irrespective of
capacity, as they do not decouple frequencies. These results reinforce
the idea that the denoising ability is a result of the frequency bias.

\subsection{Upsampling Layers}

\renewcommand{\hh}{50pt}
\renewcommand{\ww}{50pt}
\begin{wrapfigure}{r}{0.5\textwidth}
\vspace{-35pt}
\centering
\setlength{\tabcolsep}{1pt}
\begin{tabular}{ c c }
\includegraphics[width=55pt,height=\hh]{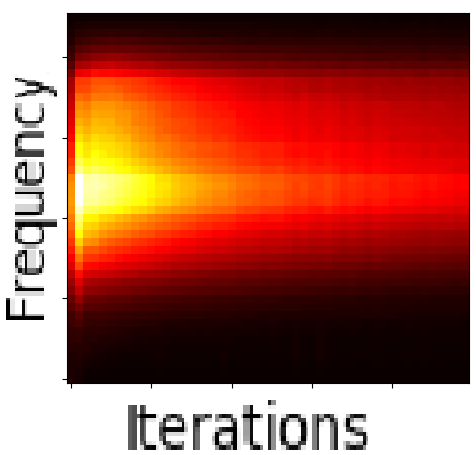} &
\includegraphics[width=\ww,height=\hh]{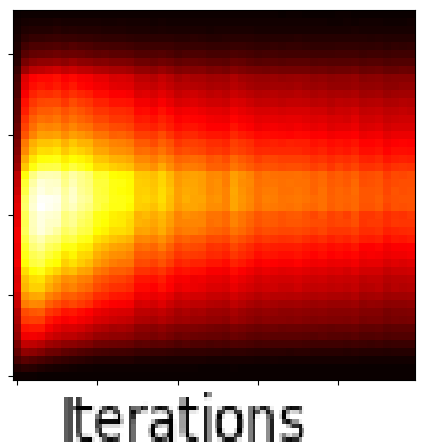} \\
(a) & (b) \\
\includegraphics[width=\ww,height=\hh]{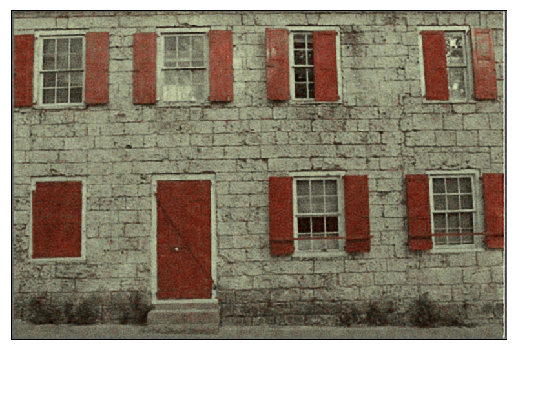} &
\includegraphics[width=\ww,height=\hh]{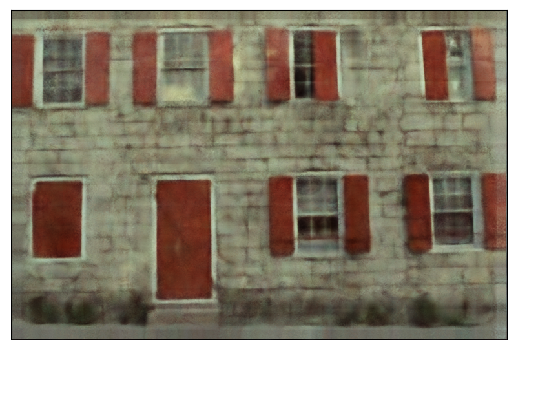} \\
(c) & (d) \\
\end{tabular}
\caption{\textbf{(a-b)} Error in the power spectrum across the first $500$ iterations using (a) stride=$4$ and (b) stride=$32$. \textbf{(c-d)} Output at $500$ iterations with (c) stride=$4$ and (d) stride=$32$. A larger stride produces a smoother output at the same iteration.}
\label{stride_effect}
\end{wrapfigure}

Two types of upsampling methods were explored in \cite{ulyanov2018deep}: nearest neighbor and bilinear.
Both methods can be viewed as upsampling with zeros followed by a
convolution with a fixed smoothing kernel. 
These strongly bias the model towards smooth images.
For example, upsampling a 1D signal with stride $L$, the frequency
responses of the associated convolution operations are ${\sin(\pi
  L k)}/{\pi L k}$ for nearest-neighbor.
and ${\sin^2(\pi L k)}/{\pi ^2 L k^2}$ for  bilinear upsampling
(Appendix \ref{app_freq_upsampling}). 
The responses decay with frequency and stride. 
This is qualitatively demonstrated in Fig.~\ref{stride_effect}. 
The low error region of the power spectrum (dark region at the bottom of
Fig.~\ref{stride_effect}(a,b)) grows more slowly, with a higher
stride.

\section{Conclusion}
This paper studies the DIP, proposing an explanation for
the denoising experiments presented in \cite{ulyanov2018deep}. We
associate the phenomenon with the notion of frequency bias:
convolutional networks fit the low frequencies of a signal
first. Experiments on 1D signals and images with additive Gaussian
noise strongly suggest that this is the causative factor behind the
denoising behavior observed in DIP.



{
\small
\bibliographystyle{ieee}
\bibliography{references}

\begin{thebibliography}{10}\itemsep=-1pt

\bibitem{arpit2017closer}
D.~Arpit, S.~Jastrzebski, N.~Ballas, D.~Krueger, E.~Bengio, M.~S. Kanwal,
  T.~Maharaj, A.~Fischer, A.~Courville, Y.~Bengio, et~al.
\newblock A closer look at memorization in deep networks.
\newblock In {\em Proceedings of the 34th International Conference on Machine
  Learning-Volume 70}, pages 233--242. JMLR. org, 2017.

\bibitem{cheng2019bayesian}
Z.~Cheng, M.~Gadelha, S.~Maji, and D.~Sheldon.
\newblock A bayesian perspective on the deep image prior.
\newblock {\em arXiv preprint arXiv:1904.07457}, 2019.

\bibitem{gunasekar2018implicit}
S.~Gunasekar, J.~D. Lee, D.~Soudry, and N.~Srebro.
\newblock Implicit bias of gradient descent on linear convolutional networks.
\newblock In {\em Advances in Neural Information Processing Systems}, pages
  9461--9471, 2018.

\bibitem{neyshabur2017exploring}
B.~Neyshabur, S.~Bhojanapalli, D.~McAllester, and N.~Srebro.
\newblock Exploring generalization in deep learning.
\newblock In {\em Advances in Neural Information Processing Systems}, pages
  5947--5956, 2017.

\bibitem{neyshabur2014search}
B.~Neyshabur, R.~Tomioka, and N.~Srebro.
\newblock In search of the real inductive bias: On the role of implicit
  regularization in deep learning.
\newblock {\em arXiv preprint arXiv:1412.6614}, 2014.

\bibitem{rahaman2018spectral}
N.~Rahaman, D.~Arpit, A.~Baratin, F.~Draxler, M.~Lin, F.~A. Hamprecht,
  Y.~Bengio, and A.~Courville.
\newblock On the spectral bias of deep neural networks.
\newblock {\em arXiv preprint arXiv:1806.08734}, 2018.

\bibitem{saxe2011random}
A.~M. Saxe, P.~W. Koh, Z.~Chen, M.~Bhand, B.~Suresh, and A.~Y. Ng.
\newblock On random weights and unsupervised feature learning.
\newblock In {\em ICML}, volume~2, page~6, 2011.

\bibitem{soudry2018implicit}
D.~Soudry, E.~Hoffer, M.~S. Nacson, S.~Gunasekar, and N.~Srebro.
\newblock The implicit bias of gradient descent on separable data.
\newblock {\em The Journal of Machine Learning Research}, 19(1):2822--2878,
  2018.

\bibitem{ulyanov2018deep}
D.~Ulyanov, A.~Vedaldi, and V.~Lempitsky.
\newblock Deep image prior.
\newblock In {\em Proceedings of the IEEE Conference on Computer Vision and
  Pattern Recognition}, pages 9446--9454, 2018.

\bibitem{xu2019frequency}
Z.-Q.~J. Xu, Y.~Zhang, T.~Luo, Y.~Xiao, and Z.~Ma.
\newblock Frequency principle: Fourier analysis sheds light on deep neural
  networks.
\newblock {\em arXiv preprint arXiv:1901.06523}, 2019.

\bibitem{xu2018training}
Z.-Q.~J. Xu, Y.~Zhang, and Y.~Xiao.
\newblock Training behavior of deep neural network in frequency domain.
\newblock {\em arXiv preprint arXiv:1807.01251}, 2018.

\bibitem{zhang2016understanding}
C.~Zhang, S.~Bengio, M.~Hardt, B.~Recht, and O.~Vinyals.
\newblock Understanding deep learning requires rethinking generalization.
\newblock {\em arXiv preprint arXiv:1611.03530}, 2016.

\end{thebibliography}
}


\clearpage
\renewcommand{\thesubsection}{\Alph{subsection}}
\section{Appendix}

\subsection{Power Spectrum in Experiment 1}
\label{app_power_spec}
\begin{figure}[h]
\centering
\setlength\tabcolsep{4pt}
\begin{tabular}{ c c }
\includegraphics[width=35pt,height=35pt]{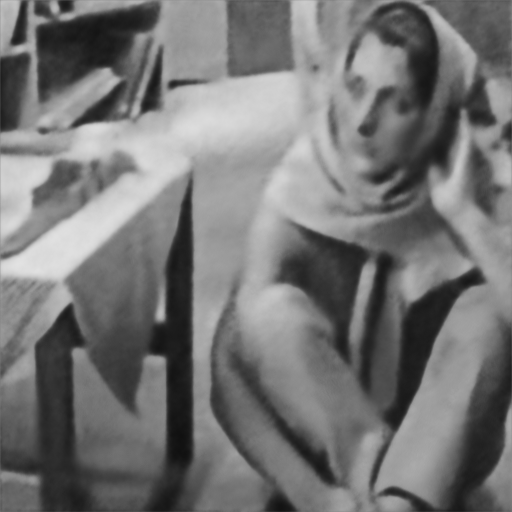}
\includegraphics[width=35pt,height=35pt]{TrajPOI/barbara_4.png}
\includegraphics[width=35pt,height=35pt]{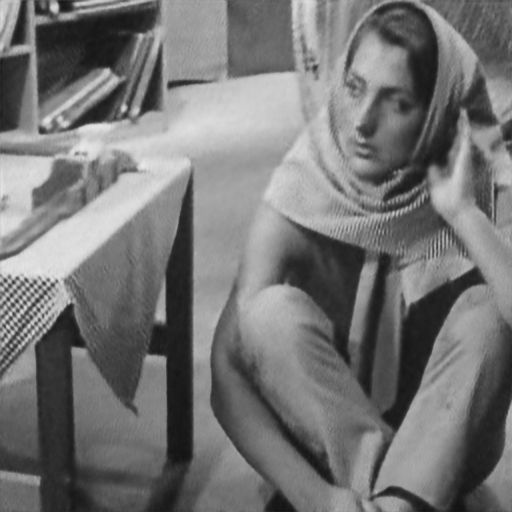}
\includegraphics[width=35pt,height=35pt]{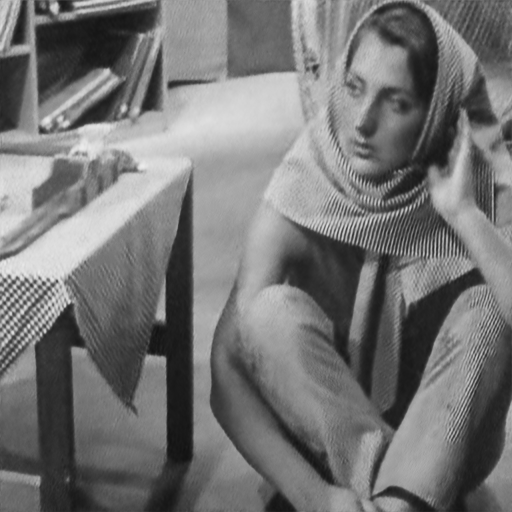}
\includegraphics[width=35pt,height=35pt]{TrajPOI/barbara_20.png} \\

\includegraphics[width=35pt,height=35pt]{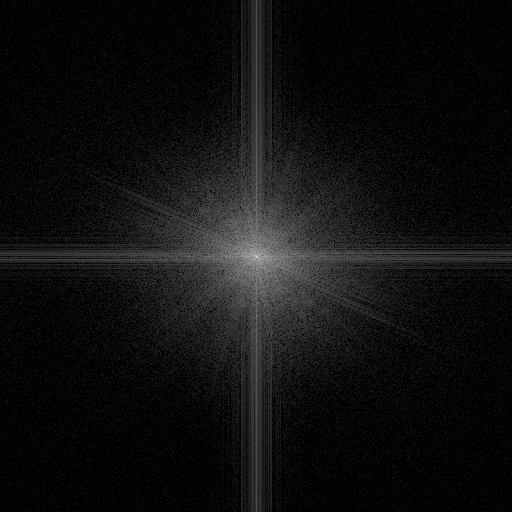}
\includegraphics[width=35pt,height=35pt]{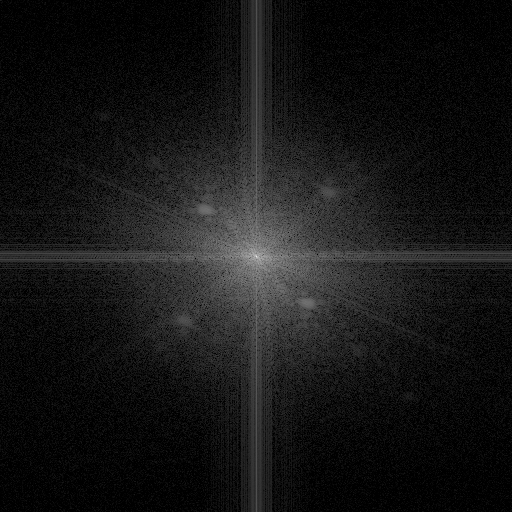}
\includegraphics[width=35pt,height=35pt]{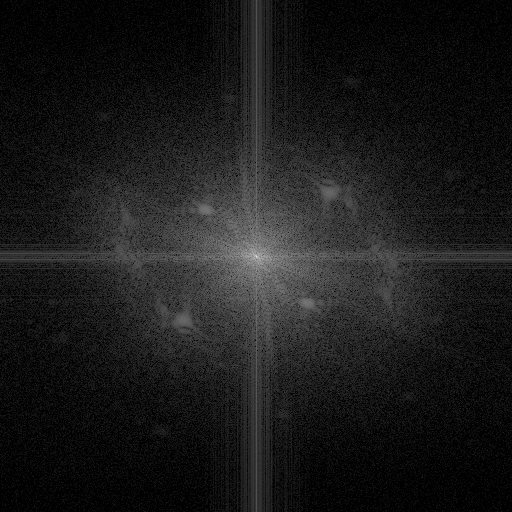}
\includegraphics[width=35pt,height=35pt]{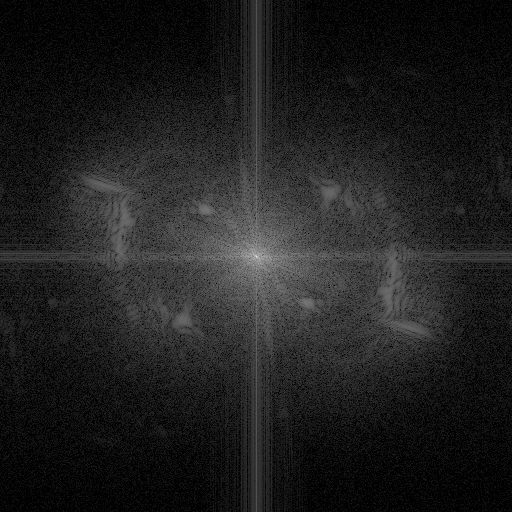}
\includegraphics[width=35pt,height=35pt]{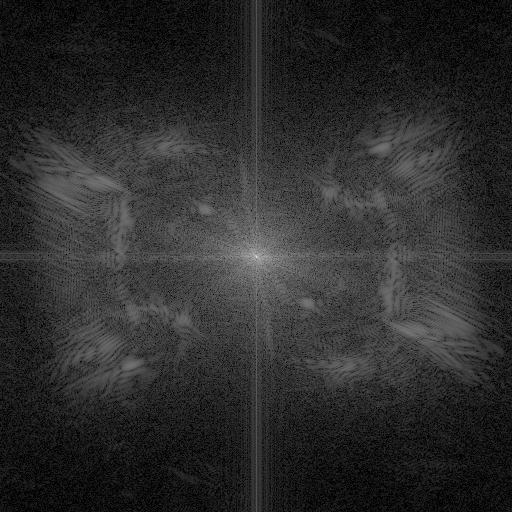} \\
\includegraphics[width=35pt,height=35pt]{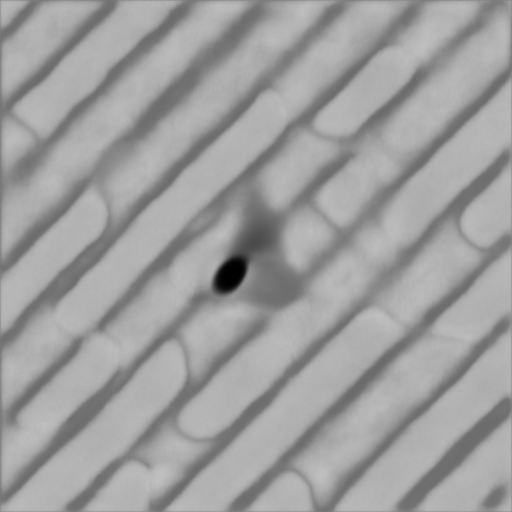}
\includegraphics[width=35pt,height=35pt]{TrajPOI/triangle_4.png}
\includegraphics[width=35pt,height=35pt]{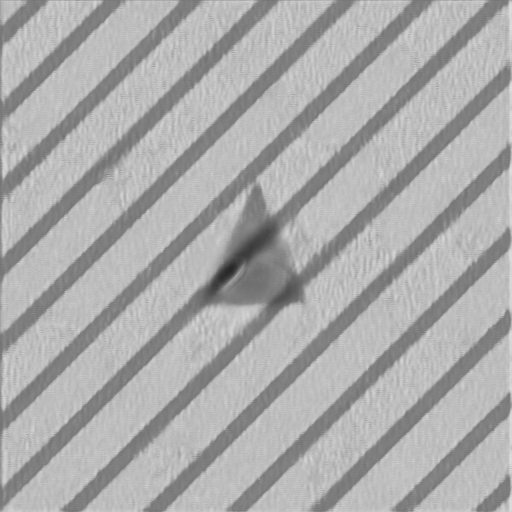}
\includegraphics[width=35pt,height=35pt]{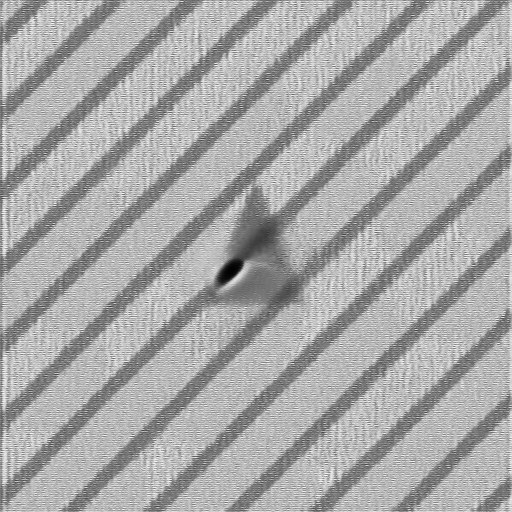}
\includegraphics[width=35pt,height=35pt]{TrajPOI/triangle_20.png} \\
\includegraphics[width=35pt,height=35pt]{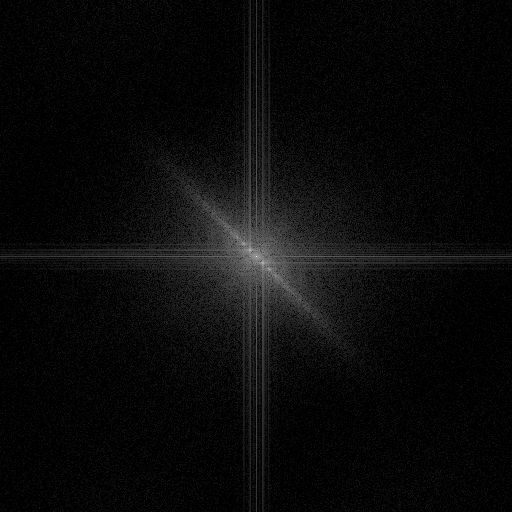}
\includegraphics[width=35pt,height=35pt]{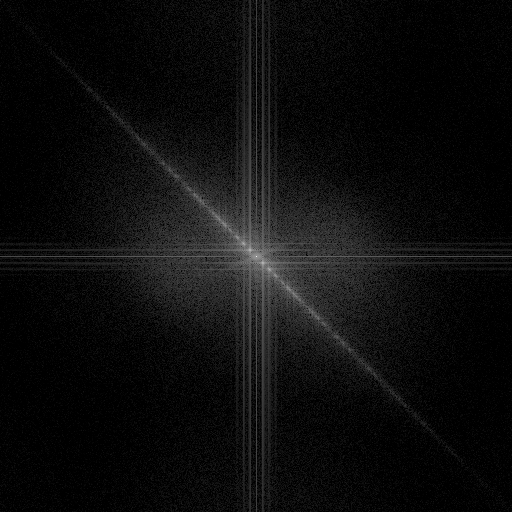}
\includegraphics[width=35pt,height=35pt]{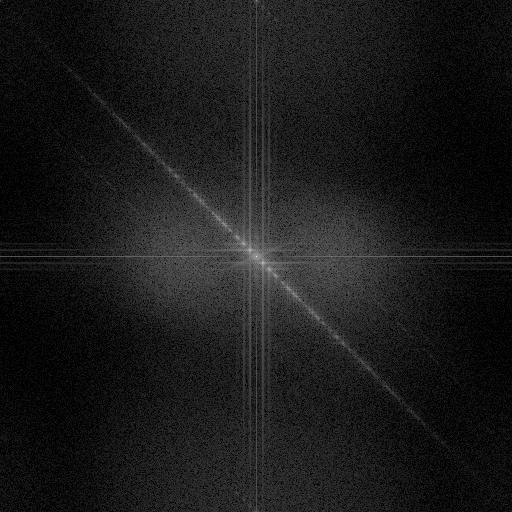}
\includegraphics[width=35pt,height=35pt]{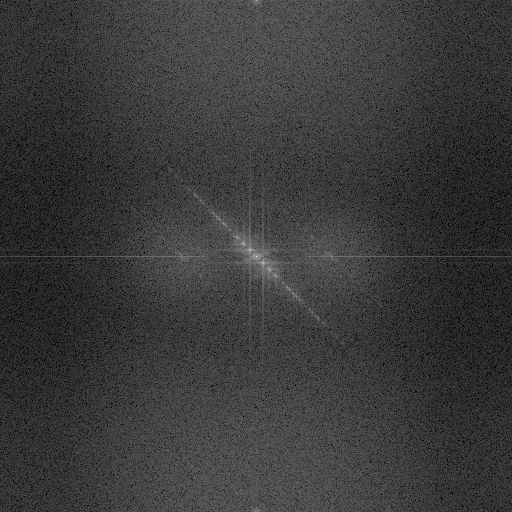}
\includegraphics[width=35pt,height=35pt]{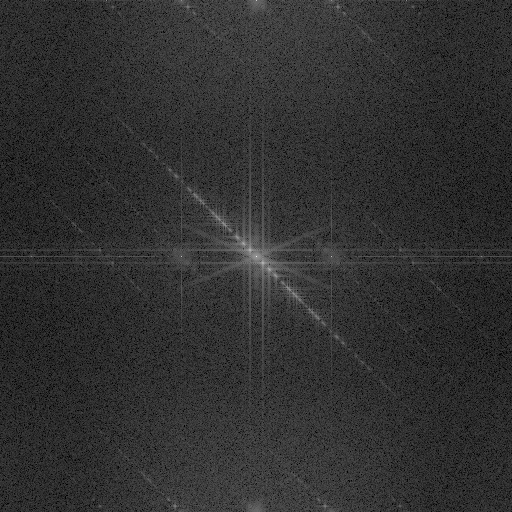} \\
\end{tabular}
\caption{\textbf{The power spectrum of DIP output on clean images.}
  Iterations $20$, $40$, $60$, $80$ and $200$ for (a) barbara.png and
  (b) triangle.png. \emph{(Please zoom in to see the patterns clearly.)}}
\label{dip_demo_fft}
\end{figure}

Fig.~\ref{dip_demo_fft} shows the trajectory in Exp. 1
Fig.~\ref{dip_demo}) and the corresponding power spectrum. This
clearly shows the frequency components growing from low to high
frequencies.

\subsection{Effect of model capacity}
We study the effect of changing the model capacity in the experiments
with 1D signals with 2 frequency components (Section 3.1). The depth
and width of the model was varied, keeping all other settings
fixed. For each capacity, we record the mean and standard deviation of the time to convergence for each frequency component across $10$ runs.

\begin{figure}[h]
\centering
\begin{tabular}{ c c c c }
\includegraphics[width=75pt,height=70pt]{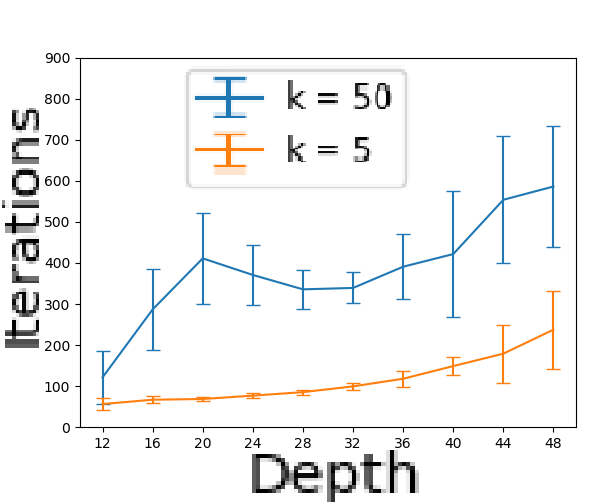} & 
\includegraphics[width=70pt,height=70pt]{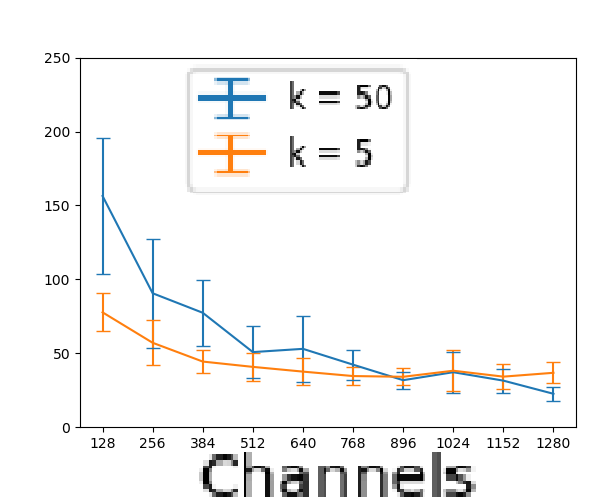} & 
\includegraphics[width=70pt,height=70pt]{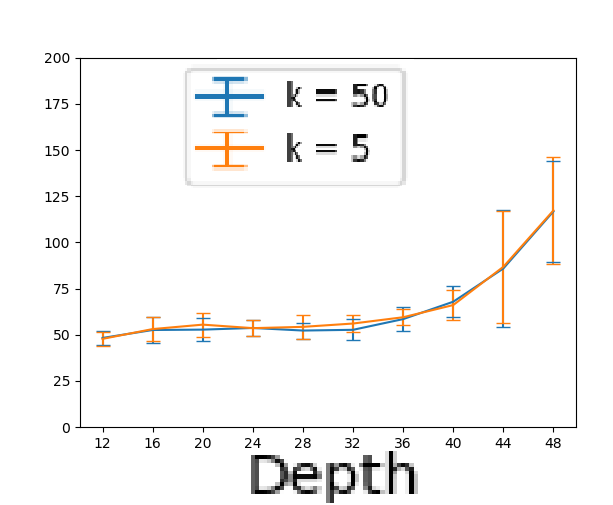} & 
\includegraphics[width=70pt,height=70pt]{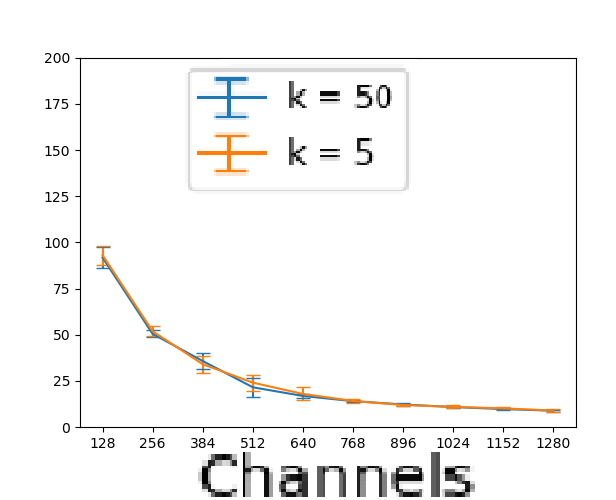} \\
(a) & (b) & (c) & (d)
\end{tabular}
\caption{Effect of varying capacity on the convergence time of frequency components $k_1$ and $k_2$. \textbf{(a-b)} Results with convolution layers: convergence time with (a) increasing depth (b) increasing number of channels per layer. \textbf{(c-d)} Results with linear layers: convergence time with (c) increasing depth (d) increasing number of nodes per hidden layer.}
\label{arch_conv}
\end{figure}

Fig.~\ref{arch_conv}(a,b) show the results when the model uses only
convolution layers. 
We observe the following:
\begin{itemize}[label=\textbullet,nosep,leftmargin=*]
    \item Increasing the depth  makes the model separate frequencies more. In Fig. \ref{arch_conv}(a), as the depth increases, the mean convergence time for $k_2$ increases much faster than that for $k_1$, indicating that the model fits the higher frequency more slowly.
    \item Increasing the width (number of channels per layer) leads to faster convergence for both frequencies (fig. \ref{arch_conv}(b)). The convergence time of $k_2$ drops faster, eventually becoming lower than that of $k_1$ ($1280$ channels per layer generally fits the higher frequency first).
\end{itemize}
These results suggest that a deep, narrow model will be more effective at decoupling frequencies than a wide, shallow one.
Fig.~\ref{arch_conv}(c,d) show the results when the model uses only
fully-connected layers. The overall trend is the same as that of the
convolutional model: time to convergence increases with depth and
decreases with width. However, both frequency components converge at
the same time. This supports the conclusion that fully-connected
layers are unable to decouple frequencies.

\subsection{Detailed Denoising Results}
\label{per_image_results}
Denoising results on each image using the architectures in section \ref{conv_layers_bias}.
Table \ref{deniose_res} summarizes the table below, showing the means per column.
\begin{table}[h]
  \centering
  \begin{tabular}{c  c  c  c  c }
  
    \hline
    \small{\textbf{Image}} & 
    \small{\textbf{DIP}} & 
    \small{\textbf{DIP Linear-128}} &
    \small{\textbf{DIP Linear-2048}} &
    \small{\textbf{ReLUNet}} \\ 
    \hline
    
    Baboon & 24.73$\pm$0.04 & 20.48$\pm$0.06 & 20.35$\pm$0.43 & 24.56$\pm$0.02 \\ 
    
    F16 & 27.00$\pm$0.15 & 20.72$\pm$0.12 & 18.60$\pm$1.85 & 26.67$\pm$0.14 \\ 
    
    House & 26.89$\pm$0.10 & 20.69$\pm$0.04 & 18.89$\pm$1.45 & 27.21$\pm$0.22 \\ 
    
    kodim01 & 25.90$\pm$0.08 & 20.35$\pm$0.07 & 20.54$\pm$0.07 & 25.95$\pm$0.07 \\ 
    
    kodim02 & 29.20$\pm$0.13 & 20.74$\pm$0.01 & 20.59$\pm$0.19 & 29.99$\pm$0.09 \\ 
    
    kodim03 & 29.26$\pm$0.13 & 20.47$\pm$0.04 & 18.13$\pm$0.13 & 29.62$\pm$0.16 \\ 
    
    kodim12 & 29.61$\pm$0.06 & 20.36$\pm$0.03 & 17.74$\pm$0.04 & 29.63$\pm$0.16 \\ 
    
    Lena & 27.49$\pm$0.06 & 20.48$\pm$0.01 & 18.67$\pm$0.72 & 27.78$\pm$0.16 \\
    
    Peppers & 27.12$\pm$0.11 & 20.60$\pm$0.01 & 18.98$\pm$1.31 & 26.82$\pm$0.16 \\
    
    \hline
  \end{tabular}
  \caption{\textbf{Per-image denoising performance}. Mean and standard deviation of the PSNR across $5$ runs.}
  \label{deniose_res_detailed}
\end{table}

\subsection{Trajectories of denoising experiment}
\label{app_traj_denoise}
\begin{figure}[h]
\centering
\begin{tabular}{ c c }
\includegraphics[width=50pt,height=50pt]{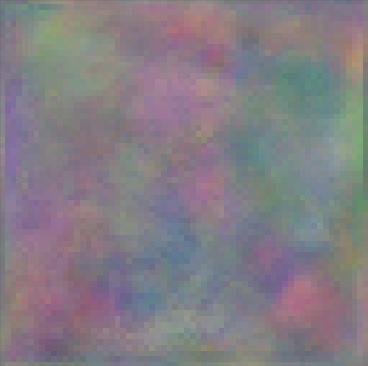} 
\includegraphics[width=50pt,height=50pt]{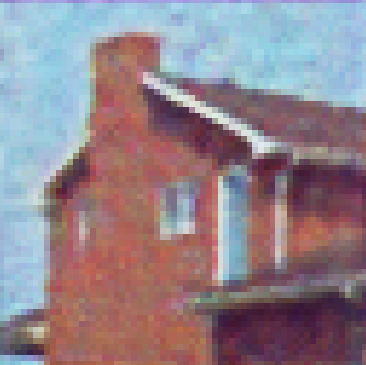}  
\includegraphics[width=50pt,height=50pt]{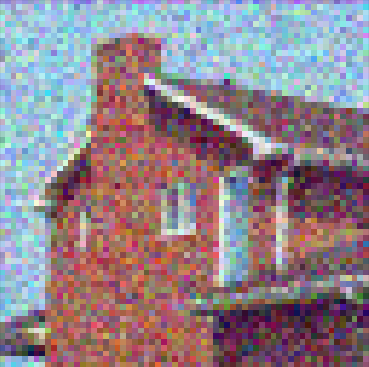} & 
\includegraphics[width=50pt,height=50pt]{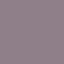}
\includegraphics[width=50pt,height=50pt]{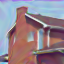}
\includegraphics[width=50pt,height=50pt]{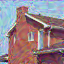} \\
(a) & (b)
\end{tabular}
\caption{Samples from trajectories when denoising House.png \textit{small}. (a) Output of DIP model at iterations $0$, $100$, $500$ (b) Output of ReLUNet model at iterations $0$, $10000$, $50000$}
\label{traj_denoise_samples}
\end{figure}

Fig. \ref{traj_denoise_samples} shows samples from the trajectories while denoising with DIP (fig. \ref{traj_denoise_samples}(a)) and with a ReLUNet (fig. \ref{traj_denoise_samples}(b)). The input image is House.png, downsampled by a factor of $4$ (to $128$x$128$) with added Gaussian noise. Both models demostrate frequency bias and predict the clean image before fitting the noise. The samples also show the difference between the DIP and ReLUNet models. ReLUNets are extremely biased towards smooth images, taking much longer to fit the higher frequencies: the DIP model starts fitting the noise at $500$ iterations, while the ReLUNet model does so at $50000$ iterations (even then, the model doesn't fit these components perfectly, predicting a blurred version of the noise).

\subsection{Frequency response of upsampling methods}
\label{app_freq_upsampling}
Standard upsampling methods can be viewed as upsampling by padding with zeros followed by a convolution with a fixed smoothing kernel $f(t)$. Let $\hat{f}(k)$ be the frequency response of this kernel (where $k$ is the frequency). Upsampling with zeros scales the image by a factor $L$, but adds high
frequency components. These components are removed after convolution
with $f(t)$ if $\hat{f}(k)$ decays with frequency. 
The spatial extent of $f(t)$ is the upsampling stride $L$. For example, in 1D, these kernels are as follows:
\begin{itemize}[label=\textbullet,nosep,leftmargin=*]
    
    \item For nearest neighbor, the kernel is the box function:
    $
        f(t) = \begin{cases}
                    1 & \text{if} t \in [\frac{-L}{2},\frac{L}{2}] \\
                    0 & \text{otherwise}
                \end{cases}
    $
    \newline
    The frequency response of this kernel is:
    $
        \hat{f}(k) = \frac{\sin(\pi L k)}{\pi L k}
    $
    \item For bilinear upsampling, the kernel is the triangle wave function:
    $
        f(t) = \begin{cases}
                    1-|t| & \text{if} |t| \leq \frac{L}{2} \\
                    0 & \text{otherwise}
                \end{cases}
    $
    \newline
    The frequency response of this kernel is:
    $
        \hat{f}(k) = \frac{\sin^2(\pi L k)}{L\pi ^2 k^2}
    $
\end{itemize}
Clearly, both favor low frequency components, with the bilinear having a stronger bias ($\propto \frac{1}{k^2}$) than nearest neighbor ($\propto \frac{1}{k}$).

\end{document}